# Continuous time recurrent neural networks: overview and application to forecasting blood glucose in the intensive care unit

Oisin Fitzgerald[a,*], Oscar Perez-Concha[a], Blanca Gallego-Luxan[a], Alejandro Metke-Jimenez[b], Lachlan Rudd[c], Louisa Jorm[a]

[a] Centre for Big Data Research in Health, Level 2, AGSM Building, UNSW Sydney, NSW 2052, Australia
[b] Australian e-Health Research Centre, Level 7, STARS Building - Surgical Treatment and Rehabilitation Service, 296 Herston Road, Herston QLD 4029, Australia
[c] Data and Analytics, eHealth NSW, 1 Reserve Road, St Leonards NSW 2065, Australia
* Correspondence: o.fitzgerald@unsw.edu.au

## Abstract

Irregularly measured time series are common in many of the applied settings in which time series modelling is a key statistical tool, including medicine. This provides challenges in model choice, often necessitating imputation or similar strategies. Continuous time autoregressive recurrent neural networks (CTRNNs) are a deep learning model that account for irregular observations through incorporating continuous evolution of the hidden states between observations. This is achieved using a neural ordinary differential equation (ODE) or neural flow layer. In this manuscript, we give an overview of these models, including the varying architectures that have been proposed to account for issues such as ongoing medical interventions. Further, we demonstrate the application of these models to probabilistic forecasting of blood glucose in a critical care setting using electronic medical record and simulated data. The experiments confirm that addition of a neural ODE or neural flow layer generally improves the performance of autoregressive recurrent neural networks in the irregular measurement setting. However, several CTRNN architecture are outperformed by an autoregressive gradient boosted tree model (Catboost), with only a long short-term memory (LSTM) and neural ODE based architecture (ODE-LSTM) achieving comparable performance on probabilistic forecasting metrics such as the continuous ranked probability score (ODE-LSTM: 0.118±0.001; Catboost: 0.118±0.001), ignorance score (0.152±0.008; 0.149±0.002) and interval score (175±1; 176±1).

# Introduction

Irregularly measured time series are common in many of the applied settings in which time series modelling is a key statistical tool, including medicine. Observational medical data such as electronic medical records (EMRs) are generally not collected or designed for research purposes, but rather for point of care and administrative efficacy [1, 2]. There may be large variation in the timing (or presence) of measurements across a patient stay and between even seemingly similar patients [2]. Despite these limitations, the detail and increasing ubiquity of EMRs have made them a core component of research focused on building data driven tools for precision/personalised medicine [1, 3]. EMRs are typically high dimensional, but sparse, with potentially complex interrelationships amongst features, making data-adaptive approaches such as deep learning a core tool of modern predictive modelling in medicine [4, 5]. Key methods that account for temporality have included recurrent neural networks (RNNs), convolutional neural networks, and attention networks [6]. Generally these methods treat time as discrete, necessitating algorithmic adaptions to account for informative temporal spacing between measurements [7]. However, recent developments in deep learning offer approaches that naturally overcome irregular time series issues through modelling a continuous time process.

*Forecasting in continuous time*

RNNs and related deep learning architectures have a long history of successful application to the modelling of sequential data [8]. At a high level, RNNs function by incorporating a hidden (or latent) state that acts as a memory of past events [9]. As new observations occur the hidden state updates to best account for new information. However, as commonly applied RNN models assume discrete time steps with non-informative temporal spacing between observations of the event of interest, with the hidden state constant between observations. Several approaches, such as neural ordinary differential equations (ODEs) and neural flows [10, 11], have been proposed in recent years that allow for a model to explicitly account for the temporal spacing between measurements in a fashion that can be flexibly learnt from the data.

*Glycaemic control in the intensive care unit*

We demonstrate the use of these methods for modelling inpatient blood glucose in the intensive care unit (ICU). Glycaemic control is an important part of ICU care [12, 13]. Upon their entry to ICU stay 9-29% of ICU patients will have elevated glucose (stress hyperglycaemia; above 180 mg/dL) and 2-6% will have excessively low blood glucose (hypoglycaemia; below 70 mg/dL) [14]. Insulin is used in around 40% of cases when stress hyperglycaemia is detected [14]. The combination of variation in the use of insulin between patients, and the common use of point-of-care testing for blood glucose can lead to large variation in the frequency and timing between blood glucose measurements in the ICU. Forecasting future blood glucose values may aid identification of patients at risk of hypo or hyper-glycaemia. Previous research has generally either discretised time [15] or altered from point of care testing (e.g. using continuous blood glucose monitoring) [16-19], limiting generalisabilty across real-world settings.

*Goal and structure of article*

We aim to give an overview of autoregressive recurrent neural networks for continuous time series modelling. We focus on applications to the longitudinal data structure typically found in medical settings and on models that produce probabilistic forecasts in real time. The article is structured as follows. Firstly, we provide an overview of neural ODEs and neural flows and how they link dynamical systems and deep learning. Secondly, we describe the general computational/mathematical form the models take, variations on this form, and additional considerations that come up when training continuous time models compared to discrete time models. Next, we demonstrate use of these models to forecast blood glucose in the ICU, with detailed description of several metrics for the evaluation of probabilistic forecasts. Finally, we conclude with a critical evaluation of the methods and future directions.

# Background

## Problem statement

Many physiological time series (e.g., blood glucose) are well suited to being modelled as stochastic differential equations, a continuous time noisy version of differential equations [20]. They could feasibly be measured arbitrarily frequently, constantly vary, and due to the difficulty of accounting for all factors that influence their evolution considered partially noise driven. In studying and modelling such time series it is often the case that the observation times $\tilde{\boldsymbol{t}}_i$ for subject $i$ can be mapped to discrete values. For example, if a continuous glucose monitor was used to sample blood glucose every 30 seconds for modelling purposes time could increment by one unit with every measurement. However, we consider the situation in which the process is measured quite irregularly, making any temporal discretisation (without the addition of imputation or related methods) of questionable meaning. The methods we describe approach this problem by modelling the distribution of the time series as a continuous-time process with jumps.

Our goal is to forecast the distribution of a continuous target variable $y_i(t)$ for each subject $i$, of $N$ total subjects, given a $K_i \times D$ dimensional patient history $\widetilde{\boldsymbol{h}}_i$. This history is observed at $K_i$ irregularly spaced timepoints specified by a vector of observation times $\tilde{\boldsymbol{t}}_i$ with maximum value $T_i$. Conditional on the history this distribution is a function of time $t > T_i$

$$p_\theta\big(y_i(t)\big|\widetilde{\boldsymbol{h}}_i\big) = p_\theta\big(y_i(t)\big|\widetilde{\boldsymbol{h}}_i = \{\boldsymbol{y_i}[\tilde{t}_i], \boldsymbol{x_i}[\tilde{t}_i], \boldsymbol{x_{0,i}}\}\big)$$

where $p_\theta$ is our forecast distribution with parameters $\boldsymbol{\theta}$, and $\widetilde{\boldsymbol{h}}_i = \{\boldsymbol{y}_i[\tilde{t}_i], \boldsymbol{x}_i[\tilde{t}_i], \boldsymbol{x}_{\boldsymbol{0},i}\}$ is the irregularly sampled patient history with the notation $\boldsymbol{y}_i[\tilde{t}_i]$ indicating concatenation of the previous observations of our target $y(t)$ for each timepoint in $\tilde{t}_i$. Additionally, $\boldsymbol{x_i}[\tilde{t}_i]$ contains observations of potentially co-varying processes $\boldsymbol{x}(t)$ (e.g., blood pressure) and $\boldsymbol{x}_{\boldsymbol{0},i}$ contains static features (e.g., patient diabetic status). As we see below in practise, we will not pass the full history at each time we wish to make a forecast but use hidden states to learn the necessary degree of dependence on the past, passing only the most recent set of observations, which is denoted as $\boldsymbol{h}_{T_i,i} = \{\boldsymbol{y}_i[T_i], \boldsymbol{x}_i[T_i]\}$. Modelling $p_{\boldsymbol{\theta}}\big(y_i(t)\big|\widetilde{\boldsymbol{h}}_i\big)$ is operationalised through modelling each element of $\boldsymbol{\theta}$ with a neural network For example, if $y_i(t)$ is assumed to follow a normal distribution we produce the following $\boldsymbol{\theta}(t, \widetilde{\boldsymbol{h}}_i) = (F_\mu(\widetilde{\boldsymbol{h}}_i, t), F_\sigma(\widetilde{\boldsymbol{h}}_i, t))$ where $F_\mu$ and $F_\sigma$ are neural networks modelling the mean and variance. For simplicity, we generally drop the subject subscript below.

## Autoregressive recurrent neural networks

The models we discuss can broadly be seen as extensions of autoregressive (with exogenous input) recurrent neural networks [6, 21] to the continuous time setting (figure 1). Autoregressive models are those which take previous observations of the outcome variable $y(t)$ (along with potentially exogenous variables) as an input for the next set of predictions. Recurrent refers to the use of the previous model output, or more commonly an internal state that is mapped to the output, as an input for the next set of predictions. As the extension from the discrete to continuous time setting are based on the use of neural ODEs and neural flows we briefly review these methods first.

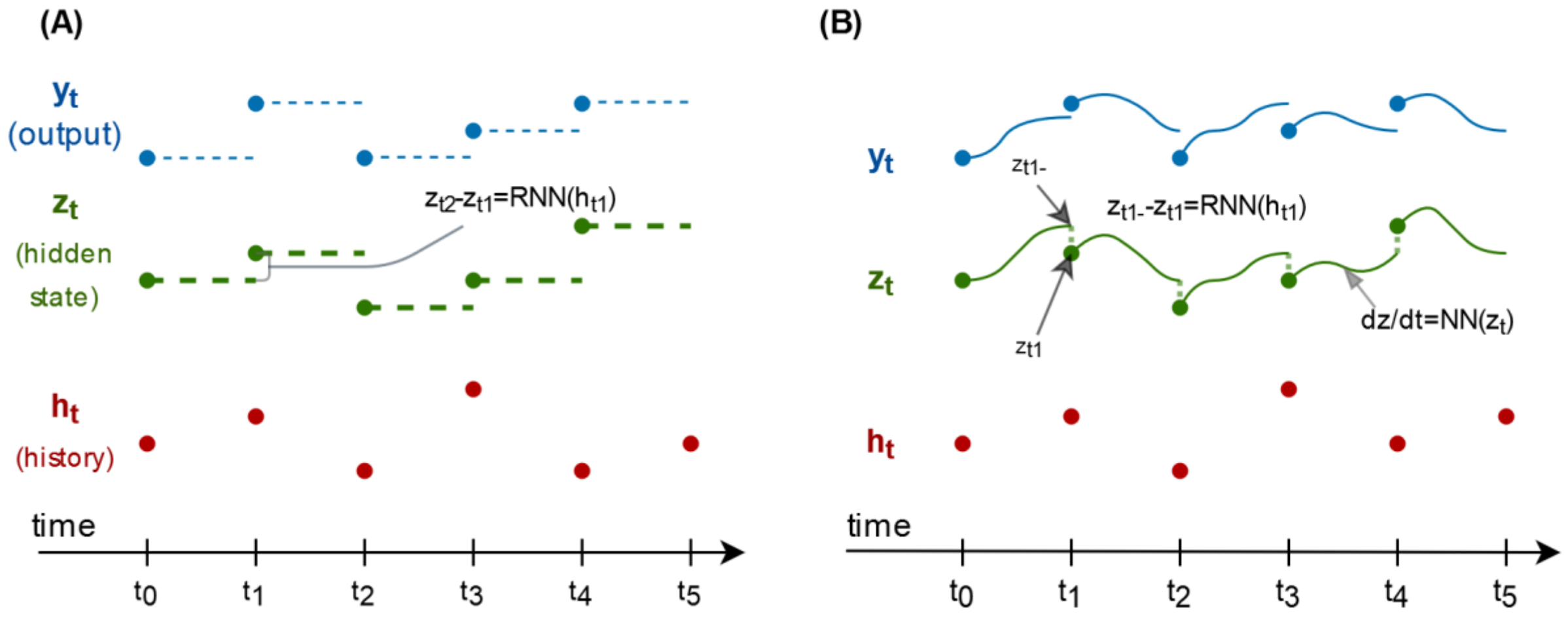

Figure 1. (A) Discrete autoregressive RNN. Discrete time RNN are a mapping $\mathbb{Z} \times \mathbb{R}^{K_i \times D} \rightarrow \mathbb{R}$ with predictions and hidden state undefined (in practise could be assumed static) between time steps. For irregularly measured time series finding an alignment (a mapping $\mathcal{A}: \boldsymbol{t}_i \rightarrow \mathbb{Z}$ may necessitate imputation or interpolation. (B) Continuous time autoregressive RNN. The hidden state $z(t)$ evolves continuously between time steps according to the neural network. The continuous time model could be seen as a fusion of the interpolation/imputation steps needed for modelling irregular time series within a single model.
Note: the notation $y_t$ is shorthand for $y(t)$

## Neural ODEs and neural flows

The models we consider have a hidden state $z(t) \in \mathcal{Z}$ that evolves continuously in time with potentially discontinuous updates when a new observation is made, as described in detail below. Between two observation times $(t_0, t_1)$ the hidden state evolves according to according to set of flow curves $F_\Theta: \mathcal{Z} \times \mathbb{R} \rightarrow \mathcal{Z}$ where $\Theta$ denotes parameters that are fixed or learnt from data. There are many approaches to parameterising this evolution, and we describe two that use neural networks below. In a neural ODEs [10] we model the vector field associated with the flow curves using a neural network $NN_{vec}$. To make a prediction using $z(t)$ at time $t_1$ we solve the initial value ODE problem.

$$\frac{dz}{dt} = NN_{vec}\big(z(t)\big), \qquad z(t_0) = z_0$$

$$z(t_1) = z_0 + \int_{t_0}^{t_1} NN_{vec}\big(z(s)\big)ds$$

in practise numerical methods are required to evaluate $z(t_1)$ , which is commonly denoted as ODESolve$(NN_{vec}, z(t), (t_0, t_1))$ (see Figure 1). The second argument to ODESolve (in this case only $z(t)$) are the inputs to $NN_{vec}$, and may contain additional information beyond simply the current value of the hidden state, as described below.

An alternative to neural ODEs, neural flows [11] use a neural network to directly parameterise the flow curves of $F_\Theta$, i.e., $z(t_1) = NN_{flow}\big(z_0, (t_0, t_1)\big)$. Careful choice of the network architecture is required to ensure the learnt neural network respects the initial conditions $NN_{flow}\big(z_0, (t, t)\big) = z_0$. While neural ODEs and flow as defined are only dependent on $z(t)$ and $(t_0, t_1)$ in both cases we may pass additional information such as the rate of ongoing drug infusions to the neural networks. We then define a set of flows $F_{x,\phi}: \mathcal{Z} \times \mathbb{R} \rightarrow \mathcal{Z}$ indexed by data $x \in X$ and $NN_{flow}$ or $NN_{flow}$ parameters $\phi \in \Phi$ which are partially fixed (e.g., number of layers) and partially learn from data (e.g., the weights).

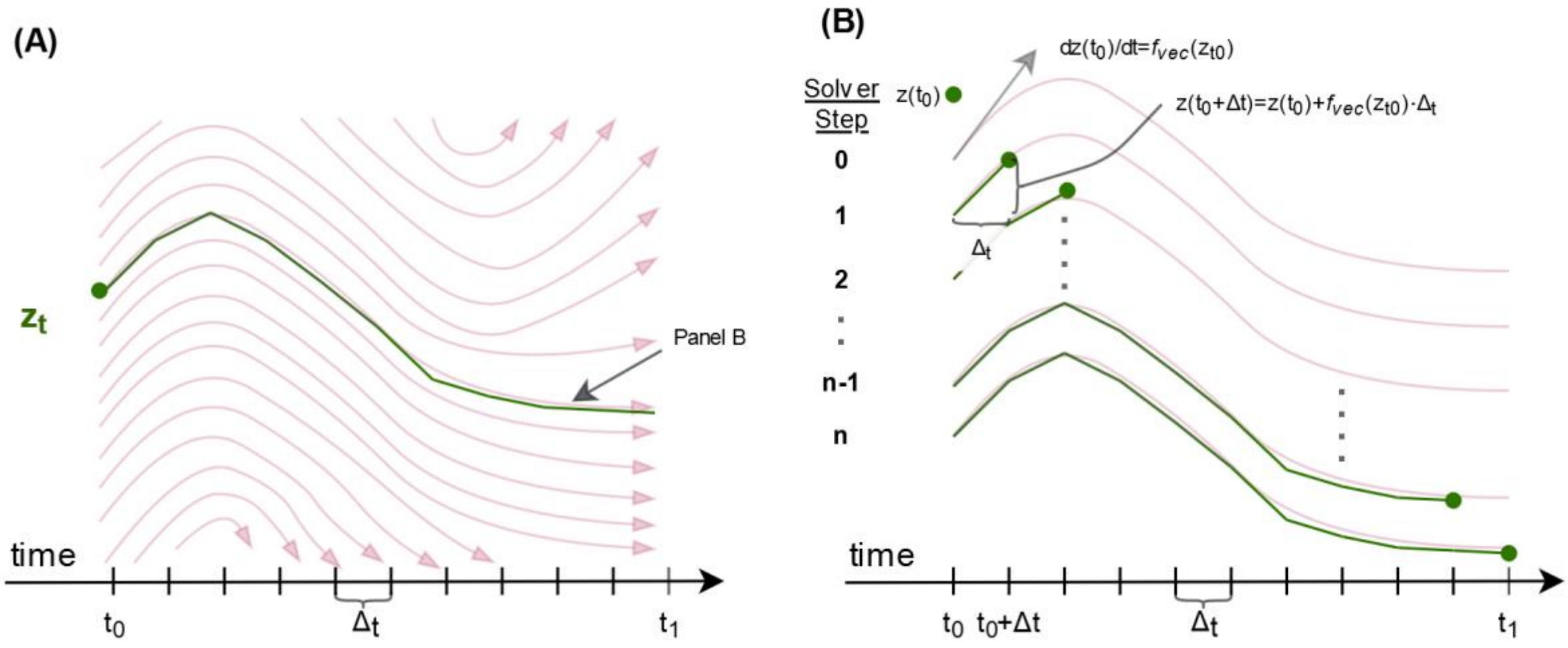

Figure 2. Illustration of process of solving (neural) ODEs (A) The flow curves defined by the neural networks $NN_{vec}$ or $NN_{flow}$ (B) Illustration of Euler method for numerically solving the ODE initial value problem - with initial state $z(t_0)$ and dynamics $dz/dt = NN_{vec}(z)$, one approach to solving neural ODEs.

# Continuous time autoregressive recurrent neural networks

Continuous time recurrent neural networks (CTRNNs) have been proposed several times [22] often for the purpose of solving differential equations. Their applicability to the analysis of medical time series has been highlighted in recent years [10, 23]. They are variations of recurrent neural networks (RNN) [24] that incorporate continuous evolution in their hidden state between observation times $(t_0, t_1)$ enabling predictions to be made at any time point. This makes them well suited to the real time prediction case that we consider. Note that it is not necessary that the flow is parameterised by a neural network, i.e., $\phi$ may define an exponential decay process [25] or set of splines [26]. With outcome distribution $P_{\boldsymbol{\theta}}$, neural networks: $NN_{init}$, $NN_{jump}$, $NN_{out}$ and flow $F_{x,\phi}$, CTRNNs follow the general forward mode algorithm.

Algorithm 1. Continuous time autoregressive recurrent neural networks

i. Initialisation:

$$z(t_0) = NN_{init}(x_0)$$

ii. Update: With new observation vector $\boldsymbol{h}_t$ set

$$z(t) = NN_{jump}(\boldsymbol{h}_t, z(t-))$$

iii. Evolve:

$$z(t + \Delta_t) = F_{x,\phi}(z(t), t + \Delta_t)$$

iv. Forecast:

$$\boldsymbol{\theta} = NN_{out}(z(t + \Delta_t))$$

$$y(t + \Delta_t) \sim P_{\boldsymbol{\theta}}$$

Note that we denote the value of the hidden state just before an update as $z(t-)$. At a high level the model consists of components that 1) perform discontinuous updates to the hidden state when a new observation is made (update); 2) propagate the hidden state $z(t)$ forward in time with a potential dependence on aspects of patient history and temporal signals (evolve); and 3) produces a probabilistic forecast as required by outputting the time dependent vector of outcome distribution model parameters $\theta(t)$ (forecast)(Figure 3). This can be represented using function composition as $\boldsymbol{\theta}((t, t+\Delta_t), \boldsymbol{h}_i) = (NN_{out} \circ F_{\mathrm{x},\phi} \circ NN_{jump})(\boldsymbol{h}_i, (t, t+\Delta_t))$.

Several variations in the architecture of continuous time autoregressive neural networks have been proposed. The distinctions lie in the architecture of the sub-networks and which inputs and hidden states are passed to the sub-networks. For example, in the ODE-RNN model proposed by [23] $NN_{vec}$ is MLP and $NN_{jump}$ is a RNN (e.g., a GRU or vanilla RNN). In contrast, the neural jump ODE [27] uses a MLP for both $NN_{vec}$ and $NN_{jump}$ with $NN_{jump}$ only depending on the input vector and not the current hidden state. In general, there are a wide variety of model architectures that fit Algorithm 1. We describe and compare some of these architectures below.

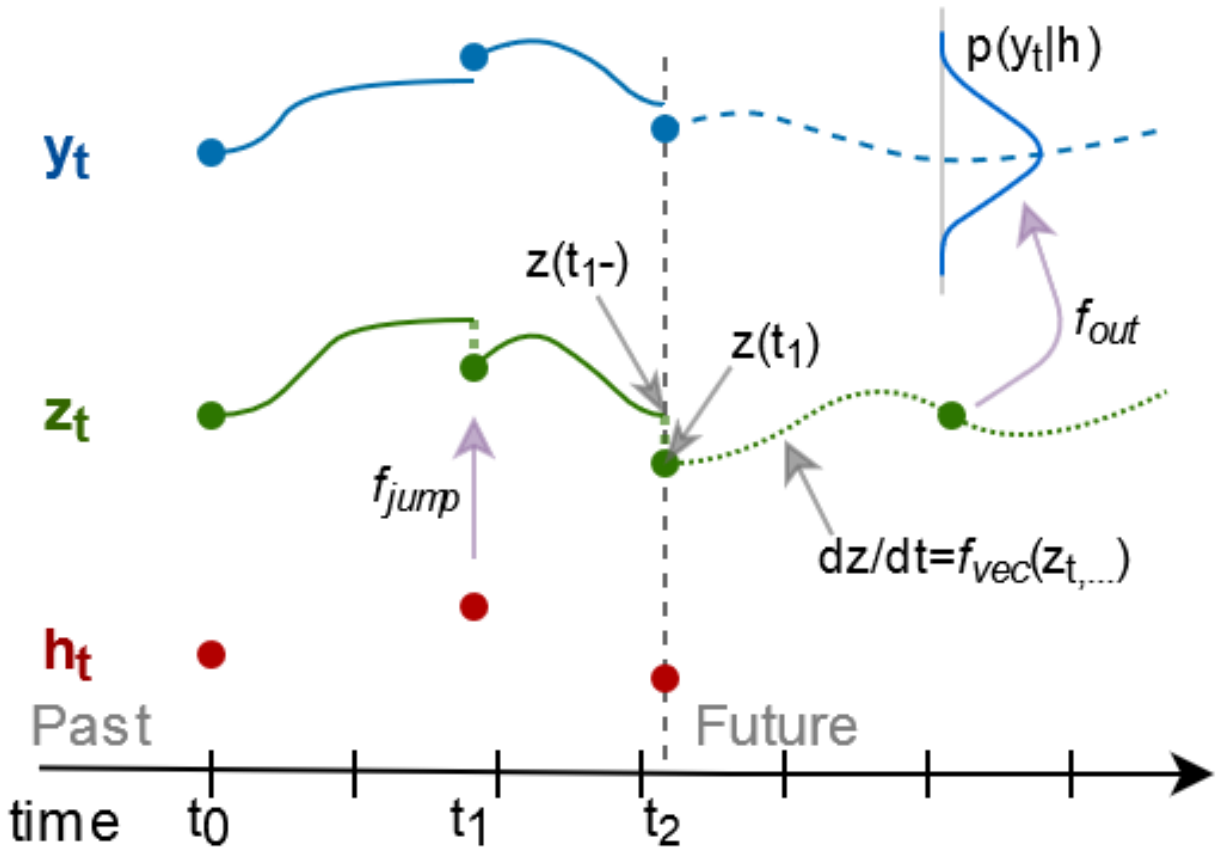

Figure 3. Illustration of model algorithm when neural ode …. When the history $h_t$ is updated with new observations $NN_{jump}$ updates the hidden state $z_t$. Between observations the state of $z_t$ evolves according to $NN_{vec}$.
Note: the notation $y_t$ is shorthand for $y(t)$

## Objective function(s)

Our model estimation largely follows [28] and [27], and can be seen as analogous to the update/filter and predict steps from the Kalman filter. As noted in [21] the distribution $P_\theta$ should match the statistical properties of the data. With an appropriate $P_\theta$ we proceed to minimising the error from the evolve/forecast steps and maximising the fit from the update step. As a result, each observation of the outcome $y(t)$ is used twice. We denote the value of the hidden state before an update as $z(t-)$ (i.e., before the data point is refed in an autoregressive fashion to the mode) and as $z(t)$ after (Figure 3). The loss function for the forecast step is the negative log likelihood of our data

$$\mathrm{Loss}_{pred}(\boldsymbol{y}_i, \boldsymbol{\theta}) = -\sum_{i=1}^{N} \sum_{t=t_{0,i}}^{T_i} \log p_{\boldsymbol{\theta}}(y|z(t-))$$

where $z(t-)$ is the final hidden state value before the update step. The additional loss function to ensure that following the update step the outcome distribution implied by the new value of the hidden state is "close" to the observed data point $y(t)$ [27, 28] is

$$\text{Loss}_{jump}(\boldsymbol{y}_i, \boldsymbol{\theta}) = -\sum_{i=1}^{N} \sum_{t=t_{0,i}}^{T_i} m_i \log p_{\boldsymbol{\theta}}(y|z(t))$$

where $m_i$ is a binary mask useful when an update step is required (e.g., due to changes in $x(t)$) but $y(t)$ is either not measured or otherwise missing. If the observation process is measured with error, we can account for this through using a Kullback-Leibler divergence as loss

$$\text{Loss}_{jump}(\boldsymbol{y}_i, \boldsymbol{\theta}) = -\sum_{i=1}^{N} \text{KL}(p_{\boldsymbol{\theta}}(y|z(t)||p_{obs}(y; \sigma^2))$$

Where $p_{obs}(y; \sigma_e^2)$ is the assumed distribution of the response $y(t)$ given observation y and an assumption of mean zero measurement error with variance $\sigma_e^2$.

This gives a final objective function with tuneable parameter $\lambda$ that accounts for potential differences in magnitudes of the losses.

$$\text{Loss} = \text{Loss}_{pred} + \lambda \, \text{Loss}_{jump}$$

## Model training

Continuous time autoregressive recurrent neural networks may be trained using standard deep learning training algorithms built on automatic differentiation packages. As noted by [29] models incorporating neural ODEs may be trained in conjunction with the rest of network via back propagation through time (BPTT) [24] or the adjoint method [10]. The adjoint method enables calculation of the derivative of the loss with respect to the parameters of $NN_{vec}$ using a new ODE – the adjoint – which can be passed to ODESolve (and run "backwards"). This has lower memory costs than BPTT due to no requirement to store intermediate states of ODESolve from the forward pass.

## Covariates and interventions

As noted above we will commonly have time varying or static covariates as inputs along with the autoregressive component. Further, some of these covariates may be best seen as interventions (such as medical treatments) designed to control the value of $y(t)$. The time-varying covariate process $\boldsymbol{x}(t)$ can be seen as a combination of $\boldsymbol{x}(t) = (\boldsymbol{x}_{input}(t), \boldsymbol{x}_{tv}(t))$ where $\boldsymbol{x}_{input}(t)$ is the interventions (such as insulin) and $\boldsymbol{x}_{tv}(t)$ is the time-varying features that are correlated with but do not control the level of $y(t)$. It has been suggested in previous research that better performance may be gained by very explicit modelling of the treatment process [30]which we explore below.

## Additional considerations

The addition of the continuous dynamic process to the autoregressive recurrent neural network model adds additional complications compared to the discrete time setting. The vector field associated with $F_{\text{x},\phi}$ should be Lipschitz continuous for it to truly describe continuous dynamics [11]. Factors that can influence this are the choice of the ODE solver and architecture of $NN_{vec}$ or $NN_{flow}$. A review of ODE solver methods is beyond the scope of this article (see [31, 32]), however we make some brief comments. Choice of the ODE solver impacts robustness and training/evaluation times. For instance, excessive discretisation can turn the solution into a discrete residual network (ResNet) which has different properties to a continuous vector field [33]. In terms of $NN_{vec}$, we generally consider cases where this is an MLP or GRU. Given that the output of a call to ODESolve is a weighted sum of calls to $NN_{vec}$, it is recommended that the output of the final layer be bounded (e.g., using a tanh non-linearity) to avoid numerical overflow or underflow. For neural flow models $NN_{flow}$ the use of a GRU or invertible ResNet as the neural flow model ensure the initial condition and continuity properties are met [11].

# Model architectures

The continuous time autoregressive recurrent neural network (Algorithm 1) permits a variety of specific architectures. Below we describe several different architectures, followed by a comparison of their performance in forecasting blood glucose.

## Neural ODE based architectures

ODE-RNNs are a CTRNN with a neural ODE evolution of the hidden state between observations. As in the discrete time setting the RNN structure may be more complex than the simple RNN update equation. Both the Gated Recurrent Unit (GRU) [34] and Long Short-Term Memory (LSTM) cells [35] have been used in the continuous time setting. Indeed, it has been shown that the empirical failure of the basic RNN to learn long term dependences also applies in the continuous time setting [29]. The LSTM has the additional hidden (cell) state $c(t)$ which [29] kept constant between updates.

The structure of $NN_{vec}$ in the ODE-RNN as proposed by previous authors [23] was a general feedforward neural network. An alternative is to view the discrete time GRU as a difference equation, and by taking the limit as the time step went to zero [28] derive a GRU based ODE.

IMODE are a class of model designed around explicit modelling of interventions (e.g., medical treatments). In contrast to the approaches described above they add an additional hidden state $z_a$ that reflects ongoing interventions (e.g., insulin infusions) (Algorithm 2). We adapt the approach in previous work [30] to remove the update independent hidden state, necessary for sparsely measured data.

Algorithm 2. IMODE update and evolve steps

ii. Update: With new observation vector $\boldsymbol{h}_t$ set

$$z_h(t) = NN_{jump}^h(\boldsymbol{h_t}, z_x)$$
$$z_a(t) = NN_{jump}^a\left(x_{input}, z_a\right)$$

iii. Evolve:

$$z_h(t + \Delta_t) = \text{ODESolve}(NN_{vec}^h, (z_x, z_z), (t, t + \Delta_t))$$
$$z_a(t + \Delta_t) = \text{ODESolve}(NN_{vec}^a, z_a, (t, t + \Delta_t))$$

We evaluate the following ODE-RNN architectures, using a MLP as $NN_{vec}$ for the LSTM and IMODE models, and a GRU as $NN_{vec}$ for the GRU model.

- ODE-GRU
- ODE-LSTM
- IMODE

### Neural flow based architectures

Flow-RNNs are a CTRNN that use a neural flow model to evole the hidden state between observations. We evaluate the following ODE-RNN architectures, using a GRU as the neural flow model.

- Flow-GRU
- Flow-LSTM

### Alternative architectures

An alternative approach to the CTRNN is to avoid the use of neural networks to parameterise $F_\Phi$. One such method is to let $\frac{dz}{dt} = -Az$ where $A \in R$, the well-known exponential decay equation with solution $z(t + \Delta_t) = z(t)e^{-A\Delta_t}$. For illustration purposes we demonstrate this method in the simulation study with a GRU based update.

- Decay-GRU

# Application to blood glucose forecasting

## Benchmark methods

Previously we described two gradient boosted tree approaches to probabilistic forecasting of blood glucose in the ICU [15]. These were based off the CatBoost library [36] and forecast either $\boldsymbol{\theta} = (\mu, \sigma)$ for an assumed $y \sim N(\mu, \sigma)$ or $\boldsymbol{\theta} = (q_{p_1}, \dots, q_{p_n})$ for quantile $q_{p_j}$. In order to account for patient history and time gaps the models were fed lagged and summarised (e.g., running mean, min and max) versions of $\widetilde{\boldsymbol{h}}_{\boldsymbol{t}}$ (the full history) and $\Delta_t$ as additional variables, creating input vector $\boldsymbol{h}_{\boldsymbol{t}}^{+} = (\boldsymbol{h}_{\boldsymbol{t}}, \Sigma(\widetilde{\boldsymbol{h}}_{\boldsymbol{t}},), \Delta_t)$. To enable direct comparison with the deep learning approaches we use the method assuming normality, given we assume log glucose to be normally distribution. As an even simpler benchmark we use a linear model to model $\mu$ and $\sigma$ using only the most recently observed information. In the linear model the target outcome for $\sigma^2$ is the squared difference between the prediction for $\mu$ and the observation $y$ [37]. Additionally, as a form of ablation study, we consider standard autoregressive RNNs with the irregular time gaps as additional input variable, giving the following benchmark methods:

- Catboost
- LinearModel
- TimeGap-GRU
- TimeGap-LSTM

## Datasets

### Simulation study

Prior work [38] proposed a nonhomogeneous Ornstein-Uhlenbeck stochastic differential equation (SDE) [20] as a simple model of blood glucose in critically ill patients. We use this model to create synthetic dataset of irregularly measured blood glucose measures, along with covariates time, insulin input and glucose input. The model has the form:

$$dG = \gamma(G - G_b) + \beta(t)m(t) + g(t) + \sqrt{2\gamma\sigma^2}dW$$

where $G$ is blood glucose in units of mg/dL, $G_b$ is "baseline" blood glucose, $\beta(t)$ is insulin sensitivity (unobserved and assumed subject constant for most scenarios), $m(t)$ is insulin input, $g(t)$ is sugar intake and $dW$ is a Wiener "white noise" process.

We simulate $N$ trajectories over a 24-hour period; for each case simultaneously simulating $G(t)$, a measurement process and a treatment decision process (which determines the value of $m(t)$) based on figure 1 from [39]. The measurement process is designed to qualitatively mimic the characteristics of real-world data (irregular measurements and subjects with higher blood glucose, and undergoing treatment being measured more often) and results in an average of 7.5 observed values over the 24-hour period.

The modelling task is to forecast the distribution of blood glucose until the next available simulated measurement, while accounting for previous blood glucose measures and nutritional and insulin inputs. The simulation studies examine how model performance varies with 1) sample size (N = 1 x $10^3$, 5 x $10^3$ and 10 x $10^3$ and 20 x $10^3$ trajectories) 2) measurement error in the observed value of $G(t)$ and 3) non-stationarity of the insulin sensitivity $\beta(t)$. Details of the full data generation algorithm and parameter settings are described in Appendix A.

### Electronic medical records

Our dataset is constructed from the publicly available MIMIC-IV database [40] using code published alongside a derived dataset of glucose-insulin measures [41]. Researchers may request access to the database via http://mimic.physionet.org. The resulting dataset comprises deidentified health data associated with 44,334 ICU stays and 772,784 blood glucose measurement. The modelling task was to forecast the distribution of blood glucose until the next available measurement in the database, while accounting for previous blood glucose measures, and nutritional and insulin inputs. Further details of the cohort and variables used are described in Appendix B.

## Models and training

For our application to forecasting of blood glucose we use the models described above. For the purposes of modelling blood glucose, we consider insulin and nutritional inputs to be interventions.

## Training and hyperparameter tuning

For the simulation studies we perform 3 runs of each simulation setting with a different random seed for each run. An external hold-out dataset of size 1,000 trajectories was used for selection of the dimension of the deep learning model hidden state, loss mixing parameter $\lambda$ and measurement error variance $\sigma_e^2$. The results from the analysis of the EMRs are generated through 3-fold cross validation on 90% of the source data. Again, a hold-out dataset (the 10% excluded from the cross-validation runs) was used for selection of the dimension of the deep learning models hidden state, mixing parameter and measurement error variance $\sigma_e^2$. For both the deep learning and gradient boosted tree models an internal (within each cross-validation run) hold-out dataset was used to prevent over-fitting and determine when the algorithm had converged.

## Model evaluation

We evaluate our probabilistic forecasts using several metrics, reporting the average of these scores across the test datasets. For continuous probabilistic forecasts it is desirable to have a scalar summary measure that achieves its minimum if the forecast distribution $P_\theta$ (or it's density $p_\theta$) matches the true distribution $P_0$, and be strictly higher for any other $G \neq P_0$. Such measures are referred to as strictly proper scoring rules [42]. They include the continuous ranked probability score and logarithmic (or ignorance) score. The continuous ranked probability score is defined as

$$\text{CRPS}(P_\theta, y_i) = \int_{-\infty}^{\infty} \left(P_\theta(y_i) - \mathbf{1}_{\{\xi \geq y_i\}}\right)^2 d\xi$$

and corresponds to the integral of the Brier score for all possible binary cut-offs [43]. If $P_\theta$ places all its probability density at a point $\hat{y}_i$ then the score is equivalent to the mean absolute error. The logarithmic score $\text{logS}(p_\theta, y_i) = -\log p_\theta(y_i)$ measures the degree to which $y_i$ was expected under assumed distribution $p$. Clearly, the mean of logS is also equal to the log-likelihood. Additionally, we assess root mean square error (RMSE) using the predicted distribution mean and the coverage and interval score from the forecasts. The interval score, for a $100(1-\alpha)\%$ prediction interval with lower bound $l$ and upper bound $u$ is defined as

$$S(l, u, y_i) = (u - l) + \frac{2}{\alpha}(l - y_i)\mathbf{1}_{\{y_i < l\}} + \frac{2}{\alpha}(y_i - u)\mathbf{1}_{\{y_i > u\}}$$

A lower score indicates better performance. For a distribution $P_\theta$ it can be shown that the distribution of the cumulative probabilities $P_\theta(y < Y)$ should be uniformly distributed [44]. In cases of overdispersion (excessive uncertainty – e.g., for a normal distribution with fixed mean the variance predictions being too large) the variance of these probabilities is less than $\frac{1}{12} = 0.83..$, whereas in cases of underdispersion it is larger than this value. Keeping with previous research [44] we refer to this metric as the variance of the probability integral transform (Var PIT). We compute the average and standard deviation of the metrics over the cross validation runs.

## Software and reproducibility

The analysis was coded in python, with pytorch [45] used for coding the models, matplotlib [46] for producing the graphs, pandas for data manipulation and summarisation [47] and properscoring for calculation of several model metrics. The code for the paper can be found at www.github.com/oizin/irregular-ts.

## Results

### Simulation study

The results of the simulation study are shown in Figure 4. All models improved in performance as the data size increased from 1,000 to 5,000 trajectories. However, the degree of improvement varied with the further increase from 5,000 to 10,000 trajectories. The Catboost model was clearly superior on the smaller dataset with some deep learning models achieving on-par or superior performance on the larger datasets (Figure 4A). The addition of measurement error led to a decrease in performance that largely maintained the model ranking from the standard setting. However, making the effect of insulin non-stationary unexpectedly lead to slight improvement in performance across models (Figure 5A). Analysis of the standard and non-stationary results indicated that this occurred due to improved performance in cases where insulin was administered in the non-stationary setting. This was a side effect of the non-stationarity leading to a larger number of training examples with insulin treatment due to the treatment lasting twice as long on average.

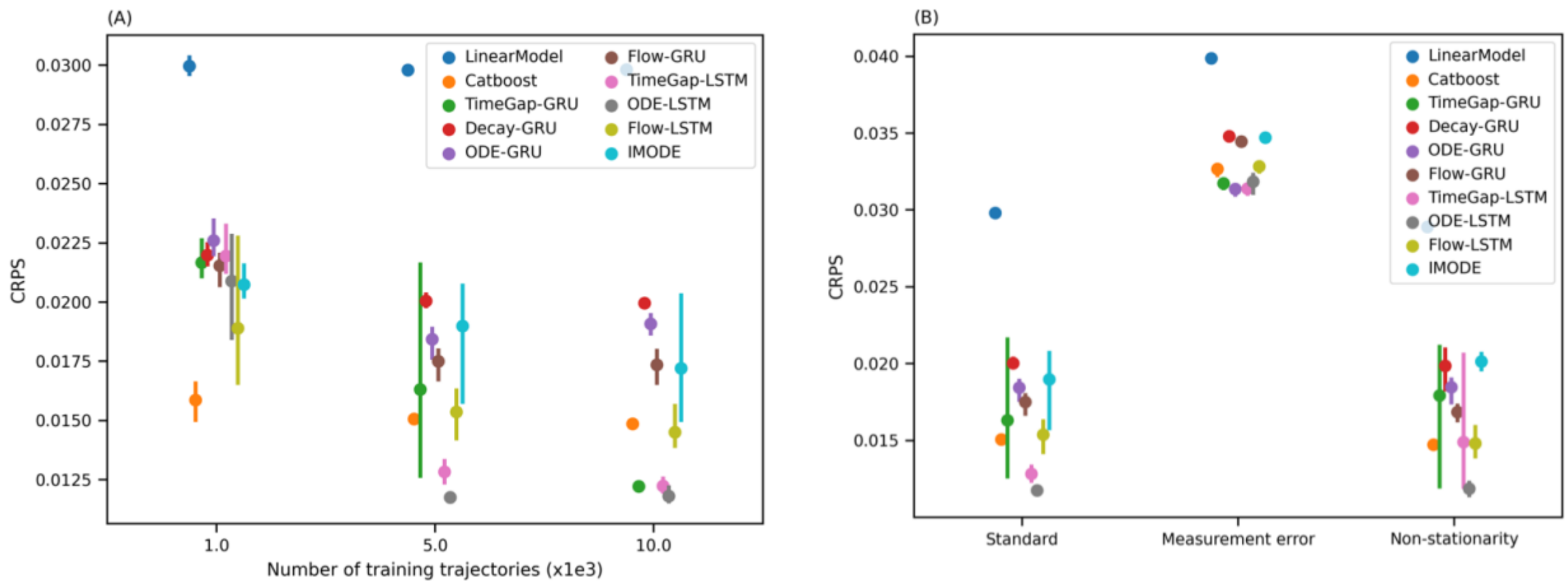

Figure 4. A) The mean, minimum and maximum of the continuous ranked probability score (CRPS) for the data size simulation B) The mean, minimum and maximum of the CRPS for the measurement error and non-stationarity simulations (N=10,000 trajectories).
Note that the number of actual data points is ~10 times the number of trajectories.

## Electronic medical records

The results of the EMR task can be seen in Table 5 (and graphically in Appendix C). The ODE-LSTM model performed best amongst the continuous time deep learning models. All continuous time RNN models outperform their discrete time analogue with added time gap variable in point prediction (RMSE), although not necessarily in accurate estimation of forecast uncertainty (e.g. examining the interval scores). Notably the Catboost model outperformed all the deep learning models except ODE-LSTM, with both of these models arguably achieving equivalent performance. As a baseline comparison using the previous observation as point prediction gives a RMSE of 45.4 mg/dL across the entire dataset (in line with the prediction from the linear model).

*Table 1. Comparison of models in on blood glucose forecasting problem, with metric mean and standard deviation across experiments*

| *Model* | *RMSE (mg/dL)* | *CRPS (x100)* | *logS (x100)* | *Var PIT (x100)* | *95% Prediction intervals* | |
|---|---|---|---|---|---|---|
| | | | | | *Empirical coverage (%)** | *Interval score* |
| *LinearModel* | 44.1±0.8 | 13.3±0.1 | -4.0±2.8 | 6.6±0.1 | 95.7±0.1 | 214±1 |
| *Catboost* | **37.1±0.2** | **11.8±0.1** | 14.9±0.2 | **7.7±0.1** | **94.3±0.2** | **176±1** |
| *TimeGap-GRU* | 39.4±0.7 | 12.4±0.1 | 10.9±0.9 | 7.6±0.1 | 94.3±0.4 | 193±2 |
| *ODE-GRU* | 39.4±0.3 | 13.0±0.2 | -23.5±6.6† | 7.3±0.1 | 93.7±0.1 | 211±5† |
| *Flow-GRU* | 38.5±0.73 | 12.3±0.1 | 12.3±2.2 | 7.5±0.1 | 94.6±0.1 | 187±1 |
| *TimeGap-LSTM* | 39.4±0.9 | 12.4±0.1 | 10.9±0.1 | 7.6±0.1 | 94.7±0.2 | 181±1 |
| *ODE-LSTM* | **36.9±0.2** | **11.8±0.1** | **15.2±0.1** | **7.6±0.1** | **94.5±0.2** | **175±1** |
| *Flow-LSTM* | 37.6±0.2 | 12.1±0.1 | 5.8±2.6 | 7.4±0.1 | 94.6±0.3 | 184±1 |
| *IMODE* | 38.5±0.1 | 12.4±0.1 | 6.0±0.5 | 8.7±0.1 | 91.2±0.2 | 198±1 |

*The interval score should be used to compare interval performance

†For 60 cases this model predicted extreme outlying values for the standard deviation.

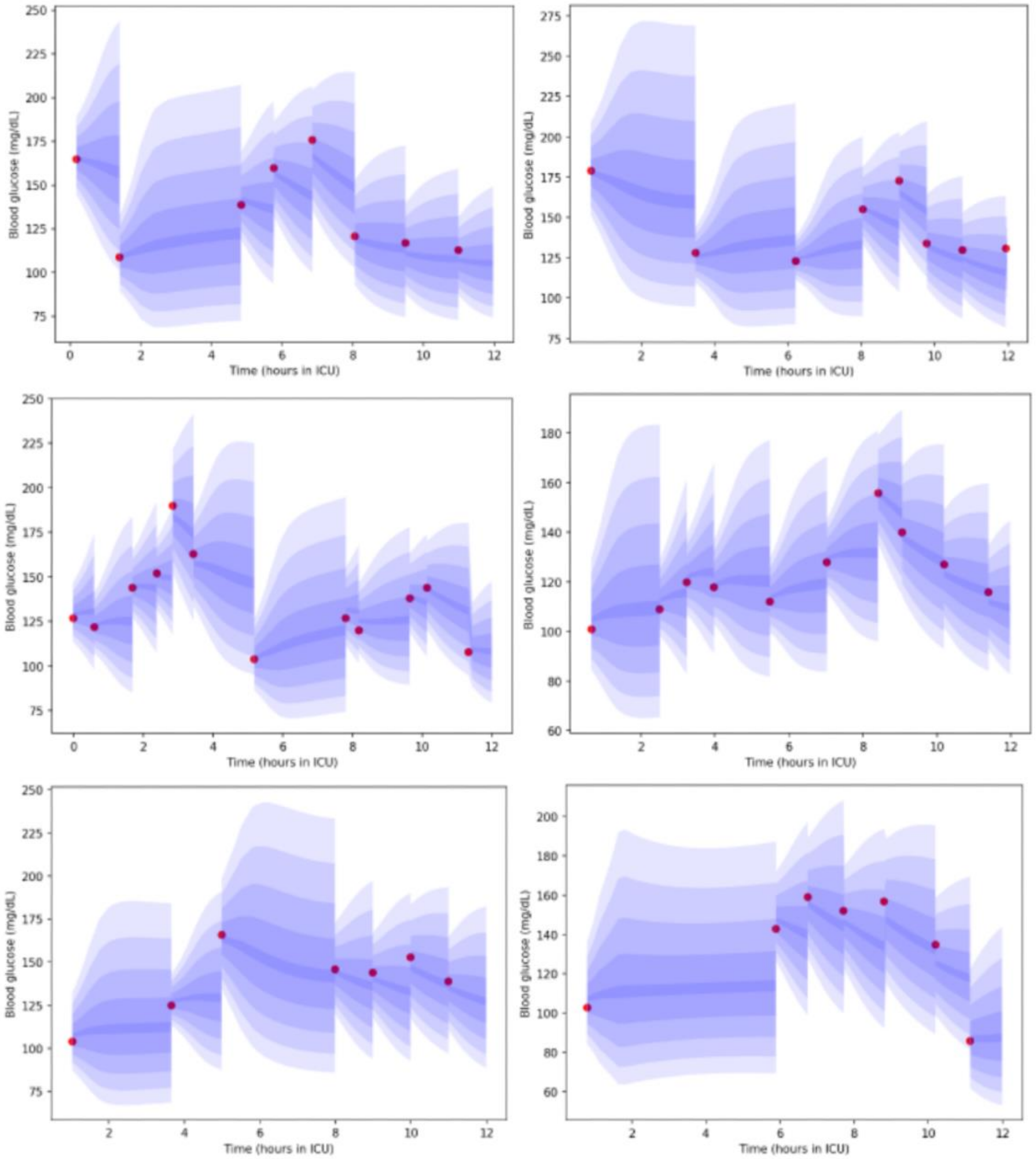

Figure 5. Example predictions from the ODE-LSTM model with 95% prediction intervals. Red dots indicate observations. Selected patients are a random subset with at least three observations within 12 hours.

# Discussion

The goal of this article is to describe and assess continuous time autoregressive recurrent neural network (CTRNN) models for probabilistic forecasting in situations with irregularly measured time series. The methods have sound theoretical basis. They build off discrete time autoregressive RNNs, incorporating a continuous time layer to enable evolution of hidden states between observations. This can account for variation in the measurement process and any continuous interventions such as medication infusions, allowing for forecasts to be made in real time. Additionally, the models are quite flexible and allow for a wide range of architecture choices depending on the problem at hand, for example the continuous intervention oriented IMODE model. Application of the methods to prediction of blood glucose in the ICU on simulated data and EMRs shows that the methods can learn from sparse irregularly measured data. Addition of a neural network based continuous time layer (neural ODE or neural flow) resulted in improved performance over a standard GRU or LSTM architecture on most measures. However, the models remain sensitive to issues such as measurement error, require sufficient data and are often outperformed by gradient boosted tree (GBT) methods.

In the EMR setting, the continuous time models produced mixed results, with the ODE-LSTM model performing best of the deep learning models on most metrics. The addition of the continuous hidden state dynamics generally improved the performance of the relevant discrete time model as seen in by comparing the performance of the continuous and time-gap versions of the GRU and LSTM models (Table 5). This was not unexpected, input features such as insulin and glucose infusions have an impact (a treatment effect) that varies with time [48]. However, simply including continuous dynamics did not lead to strong performance. Catboost outperformed all deep learning models except the ODE-LSTM and was far easier and faster to train, in line with previous results showing GBT methods can achieve superior performance on tabular data [49], including in hierarchical temporal data [50]. Further, the TimeGap-LSTM outperformed or achieved comparable performance with several continuous time GRU based models, highlighting that the assumptions encoded in both the update and evolve steps are key to achieving a well performing model. We are not aware of other research using probabilistic forecasting over irregular time intervals for blood glucose forecasting suggesting development of benchmarks for this task, as with mechanical ventilation [51], may be worthwhile.

The results of the simulation study demonstrate the importance of data size for training deep learning models. In contrast to Catboost which only improved slightly in performance as the data size went from 1,000 to 5,000 trajectories the deep learning models improved more dramatically. The simulations also highlighted the limitations of the models, providing some potential explanations for the order of magnitude difference in the error metrics for the EMR vs. simulations (aside from a general increase in complexity). The addition of both measurement error and unobserved non-stationarity resulted in altered performance. Measurement error is a known issue in blood glucose measurements in ICU [52, 53]. It is unlikely to have a static variance but vary both with blood glucose value, measurement device/source and other patient characteristics such as blood pressure. Insulin sensitivity is also proposed to vary considerably over time in critically ill patients [54]. While in the simulations a non-stationary insulin sensitivity resulted in improved performance this was due to an increased percentage of observations receiving insulin leading to more accurate performance on this subset of the data, highlighting an important consideration if these models are utilised in treatment recommendation.

A similar effect likely explains the overdispersion in the EMR setting results. In the EMR setting few models achieved the optimal probability integral transform variance of 0.83, with overdispersion the norm. This was likely due to unobserved factors – information either not recorded in the EMR or not included in this analysis – that resulted in a high level of model uncertainty for all methods.

The strength of the research was assessing on a difficult task in time series forecasting. Arguably our data were both irregular and temporally sparse. To a degree this may limit the applicability of the conclusions to more frequently measured, but still irregular data settings. For instance, the original IMODE model, while not suitable for the current task, was reported as performing strongly in a blood pressure problem [30] a far more frequently measured variable in the ICU. The limitations of the models and research also highlight several additional areas for future research. One key issue is model actionability –the degree to which the models have learnt meaningful physiologically relationships that could be used to guide treatment. Counterfactual prediction and some degree of model interpretability are likely key to such use cases [6, 55]. The current methods are a base upon which further refinement are possible. An immediate example from the current research would be development of models for dose-response curves for insulin-glucose. More generally, the continued development of machine learning and deep learning methodologies that account for the particularities of observational medical data may lead to advances in healthcare.

## Conclusion

In conclusion, the application of deep learning methods for forecasting in situations with irregularly measured time series show promise. However, appropriate benchmarking by methods such as gradient boosted tree approaches (plus feature transformation) are key in highlighting whether novel methodologies are truly state of the art in tabular data settings. Medical data is both statistically and causally challenging and accounting for measurement error, accurate quantification of uncertainty and production of counterfactual predictions are key goals for future research.

**Acknowledgements**

This work was supported by funding from eHealth NSW, UNSW Sydney and the Commonwealth Scientific and Industrial Research Organisation (CSIRO) as part of the industry PhD program.

# Appendix A: simulation study

## Data generation algorithm

The data generating algorithm featured three interrelated processes, a causal stochastic differential equation (SDE), a measurement process and a treatment decision algorithm. The data was generated from the following SDE

$$dG(t) = (\gamma(G(t) - G_b) + \beta m(t, G(s)) + g(t))dt + \sqrt{2\gamma\sigma^2}dW$$

where $G(t)$ is blood glucose in units of mg/dL, $G_b$ is "baseline" blood glucose, $m(t, G(s))$ is insulin input set based on a blood glucose measurement $G(s)$ at time $s \leq t$, $g(t)$ is sugar intake and $dW$ is a Wiener "white noise" process. We sample *N* trajectories from this model using the Euler–Maruyama method, each over a 24-hour period. For each of these trajectories we also concurrently simulate the measurement process and treatment processes. Only at measurement times can the settings of $m(t, G(s))$ be altered, mimicking a clinical visit. Starting at time zero we sample the next measurement time $t_{obs}$ using the following model

$$t_{obs} \sim LogNormal(m_{obs}, 0.2)$$

Where the value of $m_{obs}$ was based on G(t) and $m(t, G(s))$

$$\begin{cases} m_{obs} = b(m) \cdot \exp\left(\left(\frac{\mathrm{G(t)} - 120}{50}\right)^2\right) & \text{if } \mathrm{G(t)} < 80 \\ m_{obs} = b(m) \cdot \exp\left(\left(\frac{\mathrm{G(t)} - 120}{140}\right)^2\right) & \text{if } \mathrm{G(t)} \geq 80 \end{cases}$$

$$\begin{cases} b(m) = 3 & \text{if } m > 0 \\ b(m) = 5 & \text{if } m = 0 \end{cases}$$

This model is designed to qualitatively mimic patterns seen in real world data, where subjects with more extreme measurements, and undergoing treatment, are measured more frequently. The treatment algorithm was based on figure X in a review of insulin protocols [39]. In units of insulin units per hour the following treatment algorithm was used

$$\begin{cases} m\big(t, G(t)\big) = 0.0 \ if \ G(t) < 140 \\ m\big(t, G(t)\big) = 3.0 \ if \ 140 \leq G(t) < 160 \\ m\big(t, G(t)\big) = 10.0 \ if \ 160 \leq G(t) < 200 \\ m\big(t, G(t)\big) = 20.0 \ if \ G(t) \geq 200 \end{cases}$$

We use the following parameter settings and initial conditions per simulated trajectory

$$G(0) \sim N(140,20)$$
$$G_b \sim N(140,5)$$
$$\gamma \sim N(0.5,0.01)$$
$$\sigma \sim N(20,2)$$
$$\beta \sim N(50,5)$$

For the simulations with measurement error, we add $e \sim N(0, X)$ to the observed value, with this inaccurately measured value $\tilde{G}(t)$ being used in the modelling but the true value $G(t)$ being used to access algorithm performance.

The summarised data generation algorithm is shown below (table 1)

*Table 1. Data generation algorithm*

| | |
|---|---|
| | For $i$ in $1{:}N$ |
| *Initiate* | Sample $G(0)$ |
| *Sample trajectory* | While $t < 24$ |
| | If $t = t_{obs}$:<br>Set $m = m(t, G(t))$<br>Set next measurement time $t_{obs}$<br>$G(t) = G(t - dt) + dG$<br>$\tilde{G}(t) = G(t - dt) + e$ |
| *Return* | Observed process<br>Full process |

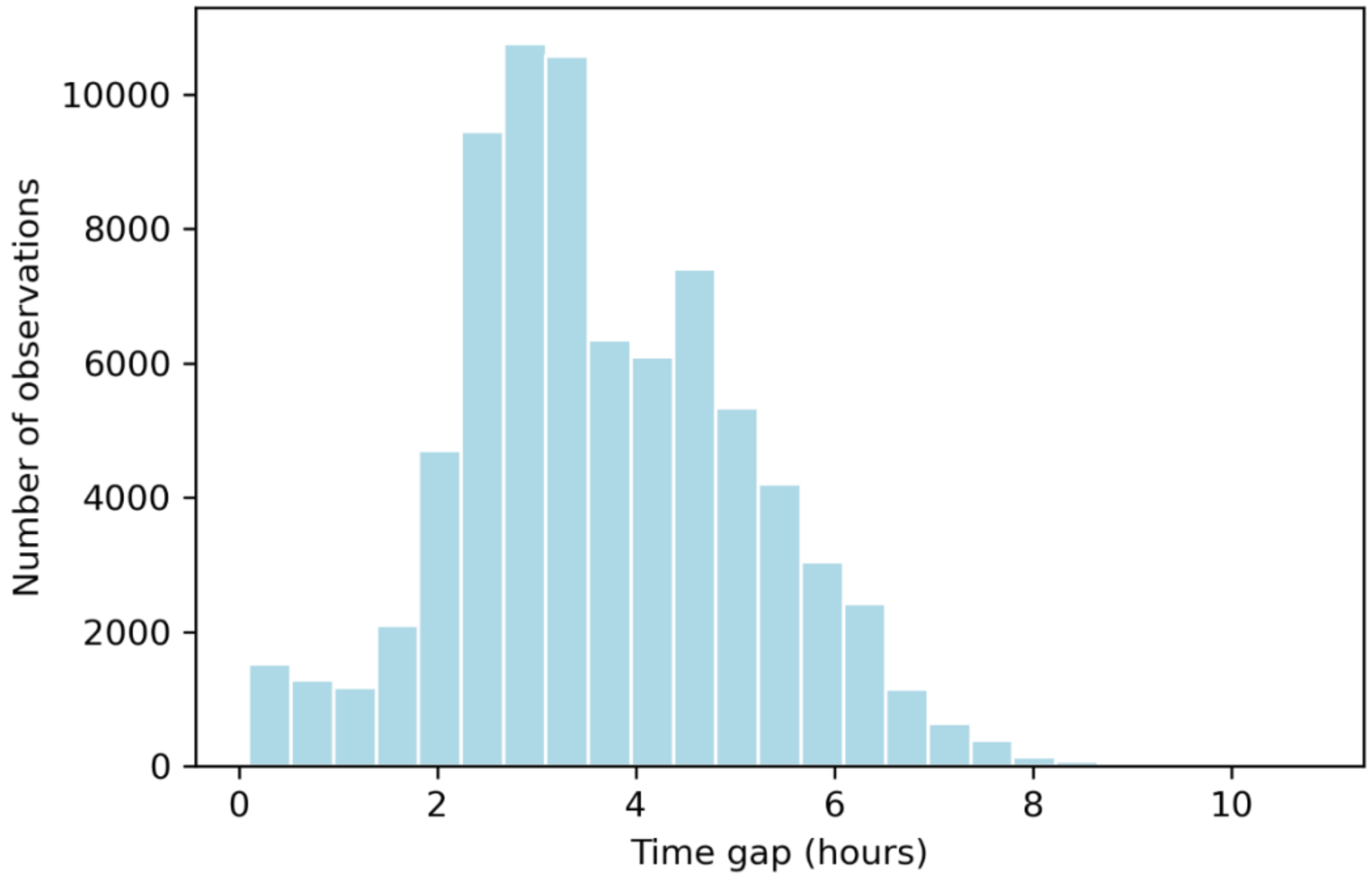

Figure A1. Distribution of time between measurements for one simulation run

# Appendix B: Electronic medical records

## Cohort

The cohort contained 772,784 blood glucose measurements across 44,334 stays.

## Variables

*Table A1. Variables used in EMR dataset*

| |
|---|
| **Outcome** |
| Blood glucose |
| **Covariates/Interventions** |
| Diabetic status (complicated/uncomplicated)<br>Age<br>Weight/Height at ICU admission<br>*Insulin*<br>Short acting (injection)<br>Short acting (bolus push)<br>Intermediate acting (injection)<br>Long acting (injection)<br>*TPN/infusions*<br>Dextrose in drug infusion (g)<br>Dextrose amount (TPN; g)<br>*Enteral feeding*<br>Enteral volume (mL)<br>CHO amount (g; enteral)<br>Dextrose amount (g; enteral)<br>Fat amount (g; enteral)<br>Protein amount (g; enteral)<br>Fibre amount (g; enteral)<br>Calorie amount (enteral) |
| **Continuous interventions** |
| *Insulin*<br>Insulin infusion |
| **Other** |
| Time to next prediction |

# Appendix C: additional results

## Simulation

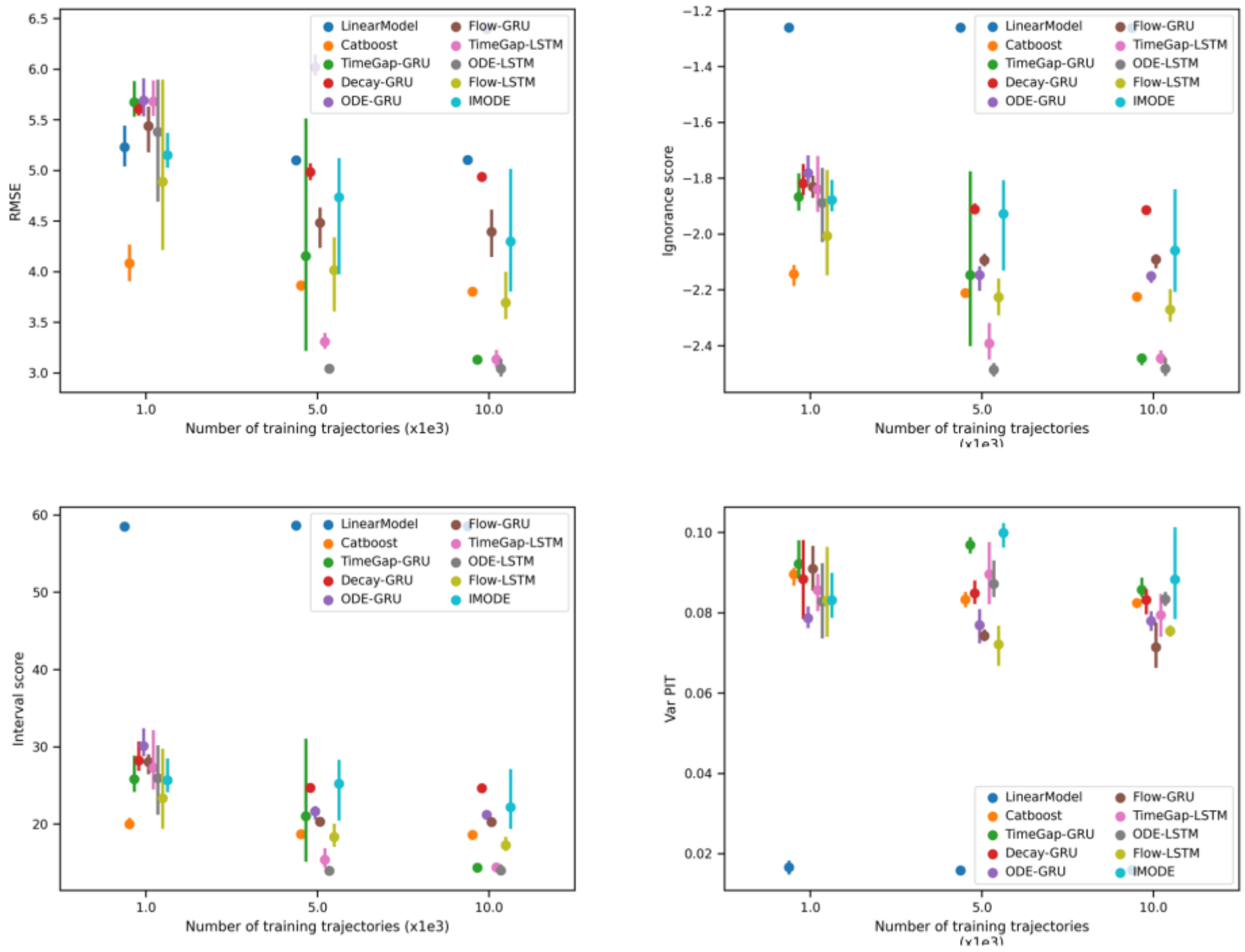

Figure C1. Mean, minimum and maximum of additional model evaluation metrics for the data size simulation.

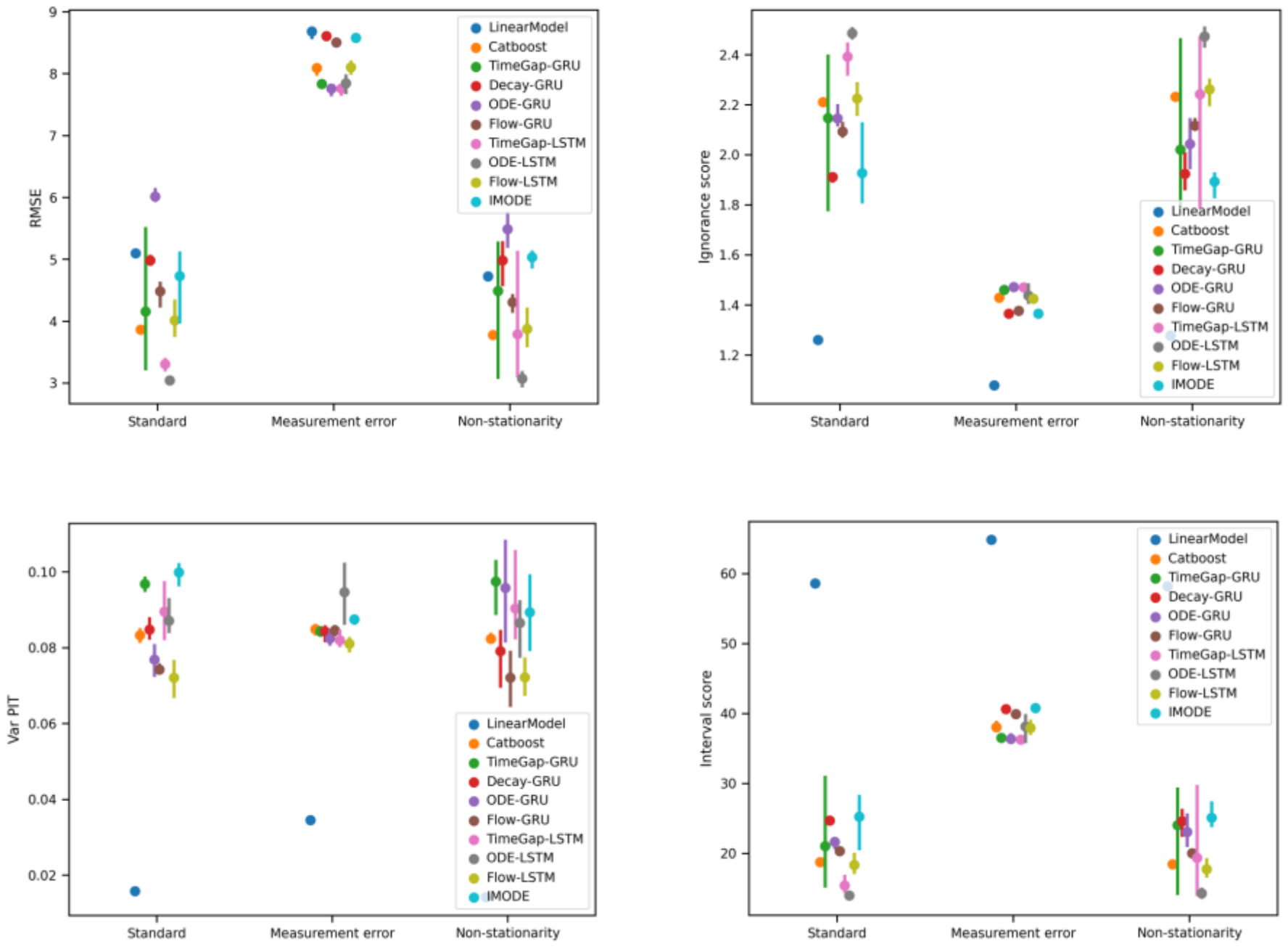

Figure C2. Mean, minimum and maximum of additional model evaluation metrics for the measurement error and non-stationary simulations.

## Electronic medical records

Below (Figure C3) we display graphically the information in Table 1 of the main manuscript.

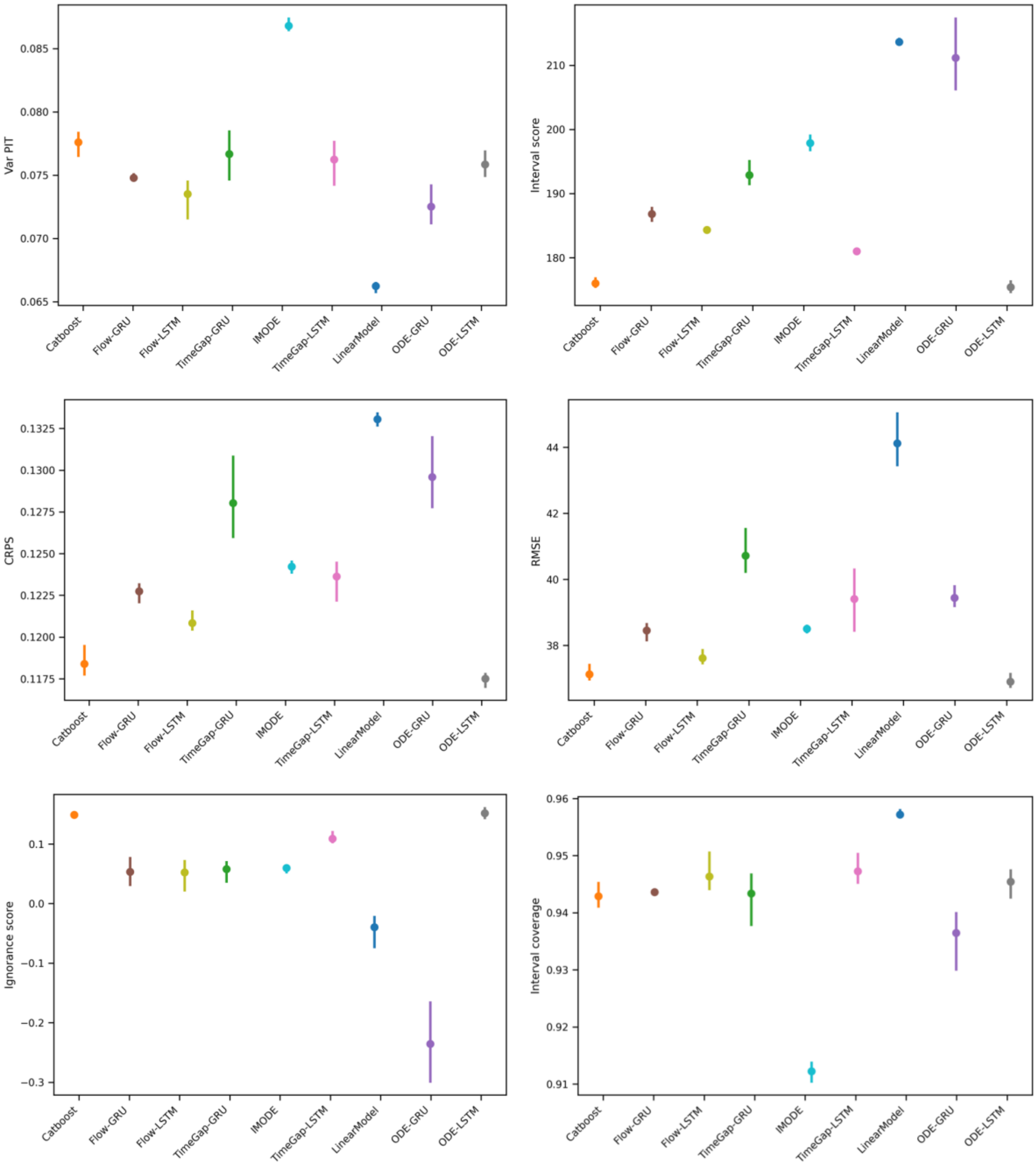

Figure C3. Mean, minimum and maximum for the model evaluation metrics for the electronic medical record dataset.